\documentclass{article}

\usepackage[numbers]{natbib}
\usepackage[preprint]{neurips_2025_custom}

\usepackage{amsmath,amsfonts,bm}

\def\eqref#1{equation~(\ref{#1})}
\def\Eqref#1{Equation~(\ref{#1})}
\def\1{\bm{1}}

\DeclareMathAlphabet{\mathsfit}{\encodingdefault}{\sfdefault}{m}{sl}
\SetMathAlphabet{\mathsfit}{bold}{\encodingdefault}{\sfdefault}{bx}{n}

\usepackage[dvipsnames]{xcolor}         % colors
\definecolor{linkColor}{rgb}{0.18,0.39,0.62}
\usepackage[utf8]{inputenc} % allow utf-8 input
\usepackage[T1]{fontenc}    % use 8-bit T1 fonts
\usepackage[colorlinks=true,linkcolor=linkColor,citecolor=linkColor,filecolor=linkColor,urlcolor=linkColor]{hyperref}       % \usepackage{url}
\usepackage{cleveref}
\usepackage{multirow}
\usepackage{xspace}
\usepackage{booktabs}
\usepackage{graphicx}
\usepackage{array}
\usepackage{wrapfig}
\usepackage{bm}
\usepackage{algorithm}

\usepackage{algpseudocode}
\usepackage{algpseudocode}
\usepackage{enumitem}
\usepackage{tcolorbox}
\usepackage{makecell}
\usepackage{diagbox}

\usepackage{amsmath}
\usepackage{amssymb}
\usepackage{mathtools}
\usepackage{amsthm}

\usepackage{multirow}
\usepackage{amsmath}
\usepackage{capt-of}
\usepackage{tabularx}
\usepackage{epsfig}
\usepackage{amssymb}
\usepackage{amsfonts}
\usepackage{booktabs}
\usepackage{scalerel}
\usepackage{listings}
\usepackage{varwidth}
\usepackage{stmaryrd}
\usepackage{bbm}
\usepackage{wrapfig}
\usepackage{pifont}

\newcommand{\tabincell}[2]{\begin{tabular}{@{}#1@{}}#2\end{tabular}}

\definecolor{deepblue}{rgb}{0,0,0.5}
\definecolor{officeblue}{RGB}{0,102,204}
\definecolor{deepred}{rgb}{0.6,0,0}
\definecolor{deepgreen}{rgb}{0,0.5,0}
\definecolor{mybrickred}{RGB}{182,50,28}

\definecolor{fillcolor}{RGB}{216,217,252}

\usepackage{etoolbox}
\usepackage{framed}

\newif\ifxetexorluatex
\ifxetex
  \xetexorluatextrue
\else
  \ifluatex
    \xetexorluatextrue
  \else
    \xetexorluatexfalse
  \fi
\fi
\newcommand*\quotesize{60} % if quote size changes, need a way to make shifts relative
\newcommand*{\openquote}
   {\tikz[remember picture,overlay,xshift=-4ex,yshift=-2.5ex]
   \node (OQ) {\fontsize{\quotesize}{\quotesize}\selectfont``};\kern0pt}

\newcommand*{\closequote}[1]
  {\tikz[remember picture,overlay,xshift=4ex,yshift={#1}]
   \node (CQ) {\fontsize{\quotesize}{\quotesize}\selectfont''};}

\colorlet{shadecolor}{white}

\newcommand*\shadedauthorformat{\emph} % define format for the author argument

\newcommand*\authoralign[1]{%
  \if#1l
    \def\authorfill{}\def\quotefill{\hfill}
  \else
    \if#1r
      \def\authorfill{\hfill}\def\quotefill{}
    \else
      \if#1c
        \gdef\authorfill{\hfill}\def\quotefill{\hfill}
      \else\typeout{Invalid option}
      \fi
    \fi
  \fi}
{\authoralign{#1}
\ifblank{#2}
   {\def\shadequoteauthor{}\def\yshift{-2ex}\def\quotefill{\hfill}}
   {\def\shadequoteauthor{\par\authorfill\shadedauthorformat{#2}}\def\yshift{2ex}}
\begin{snugshade}\begin{quote}\openquote}
{\shadequoteauthor\quotefill\closequote{\yshift}\end{quote}\end{snugshade}}

\newcommand{\github}{\raisebox{-1.5pt}{\includegraphics[height=1.05em]{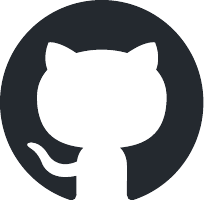}}\xspace}

\definecolor{DarkBlue}{RGB}{0, 51, 153}
\newcommand{\ours}{OPCD}

\title{On-Policy Context Distillation}
\title{On-Policy Context Distillation for \\ Knowledge Internalization}
\title{On-Policy Context Distillation for Language Models}

\author{%
Tianzhu Ye\thanks{~Equal contribution.}~~~~~~~~Li Dong\footnotemark[1] \\
\bf Xun Wu~~~~~~Shaohan Huang~~~~~~Furu Wei \\
~Microsoft Research \\
~{\href{https://aka.ms/GeneralAI}{https://aka.ms/GeneralAI}}
}

\begin{document}

\maketitle

\begin{abstract}
Context distillation enables language models to internalize in-context knowledge into their parameters. In our work, we propose \textbf{On-Policy Context Distillation} (OPCD), a framework that bridges on-policy distillation with context distillation by training a student model on its own generated trajectories while minimizing reverse Kullback-Leibler divergence against a context-conditioned teacher. We demonstrate the effectiveness of OPCD on two important applications: experiential knowledge distillation, where models extract and consolidate transferable knowledge from their historical solution traces, and system prompt distillation, where models internalize beneficial behaviors encoded in optimized prompts. Across mathematical reasoning, text-based games, and domain-specific tasks, OPCD consistently outperforms baseline methods, achieving higher task accuracy while better preserving out-of-distribution capabilities. We further show that OPCD enables effective cross-size distillation, where smaller student models can internalize experiential knowledge from larger teachers.
% , whereas directly injecting such knowledge into smaller model contexts degrades performance.
\begin{table}[H]
\centering
\begin{tabular}{@{}r@{\hspace{2pt}}l@{}}
\github & \textbf{Code}: \href{https://aka.ms/opcd-code}{\texttt{aka.ms/opcd-code}}
\end{tabular}
\end{table}
\end{abstract}

\section{Introduction}
\label{sec:intro}

Large language models (LLMs) exhibit remarkable in-context learning capabilities, allowing them to adapt their behavior based on the information provided in the prompt without parameter updates \citep{gpt3,icl:survey}.
By prepending instructions, few-shot demonstrations, or retrieved documents to the input, users can steer model behavior without updating parameters.
However, in-context knowledge is transient. In other words, valuable insights generated or retrieved during a session are lost once the context is reset, requiring the model to ``re-learn'' from the prompt every time.

A natural question arises: \textit{Can we internalize transient in-context knowledge into the model's permanent parameters?}
Context distillation \citep{context:distill:anthropic,context:distill:berkeley} addresses this by training a student model to mimic the behavior of a context-conditioned teacher, effectively compressing the context into the student's weights. Once trained, the student can reproduce the teacher's context-aware behavior without requiring the context at inference time, effectively ``internalizing'' the context.

Despite its appeal, existing context distillation methods face a fundamental limitation: they rely on off-policy training with forward Kullback-Leibler (KL) divergence minimization on a fixed dataset. However, this off-policy approach suffers from distinct drawbacks.
First, it induces exposure bias, where the student is trained on teacher-generated or ground-truth data but must generate its own autoregressive sequences at inference time.
Second, minimizing forward KL encourages mode-covering behavior, causing the student to assign probability mass to all teacher-generated tokens, often resulting in ``hallucinations'' or overly broad distributions when the student lacks the capacity to fully model the teacher's complex, context-aware distribution \citep{minillm}.

In this work, we propose \textbf{On-Policy Context Distillation} (\ours{}), a method that bridges on-policy distillation \citep{minillm,thinkingmachine-onpolicy,googlepolicy} with context distillation to internalize in-context knowledge more effectively. The key is that the student model learns from its own generation trajectories rather than those of the teacher.
Specifically, OPCD samples responses from the student model (without context), then computes the reverse KL divergence between the student's token distributions and those of a context-conditioned teacher at each position along the student's trajectory. This on-policy approach ensures that the student learns to correct its own mistakes and align its generation distribution with the teacher's context-aware behavior.

We demonstrate the effectiveness of OPCD on two important applications. First, we introduce experiential knowledge distillation, where a model extracts transferable knowledge from its historical solution traces and internalizes this accumulated experience into its parameters. We show that models can progressively improve by accumulating experiential knowledge from solved problems, and that \ours{} successfully consolidates this knowledge without requiring the extended context at inference time. Second, we apply \ours{} to system prompt distillation, enabling models to internalize beneficial behaviors encoded in externally optimized prompts for specialized tasks such as medical question answering and safety classification.

Our experiments span mathematical reasoning, text-based games, and domain-specific tasks with optimized system prompts. Across all settings, \ours{} consistently outperforms baseline methods, achieving higher task accuracy while better preserving out-of-distribution capabilities and relieving catastrophic forgetting. We further demonstrate that \ours{} enables effective teacher-student distillation, where smaller student models can internalize experiential knowledge from larger teachers. In contrast, directly injecting teacher-generated knowledge into smaller model contexts degrades performance.

% Our contributions are summarized as follows:
% \begin{itemize}[leftmargin=0pt]
% \item We propose On-Policy Context Distillation (\ours{}), a method that combines on-policy sampling with reverse KL minimization to internalize in-context knowledge into model parameters.
% \item We conduct comprehensive experiments on mathematical reasoning, text-based games, and system prompt distillation, showing that \ours{} consistently outperforms baselines while maintaining strong out-of-distribution performance.
% \item We introduce experiential knowledge distillation, a novel task where models extract, accumulate, and consolidate transferable knowledge from their own trajectories, demonstrating that accumulated experience can be effectively internalized via \ours{}.
% \end{itemize}

\section{Related Work}

\paragraph{Context Distillation}
Context distillation compresses in-context knowledge into model parameters, eliminating the inference overhead of context processing \citep{context:distill:anthropic,context:distill:berkeley,infiniteicl}. While prior methods rely on off-policy forward KL minimization, they suffer from exposure bias due to the mismatch between teacher-guided training and autoregressive inference. In contrast, our method employs on-policy sampling, allowing the student to learn from its own trajectories and bridging the gap between training and deployment distributions.

\paragraph{On-Policy Distillation}
On-policy distillation methods \citep{minillm,thinkingmachine-onpolicy,googlepolicy} mitigate exposure bias by training students on their own generated trajectories. By minimizing the reverse KL divergence~\citep{minillm}, these approaches promote mode-seeking behavior, compelling the student to focus on the teacher's high-likelihood regions and avoiding the mode-averaging issues of standard forward KL.
\citep{gad} has extended this to black-box settings.
Our work adapts the on-policy distillation paradigm specifically for the problem of context internalization, allowing a model to efficiently consolidate transient in-context knowledge into its permanent weights.

\paragraph{Self-Distillation}
Recent research has increasingly explored self-distillation mechanisms in which a model improves by learning from its own output or a conditioned version of itself.
\citep{star} demonstrates that a model can bootstrap its reasoning capabilities by iteratively training self-generated solutions that lead to correct answers.
Closer to our approach, concurrent works~\citep{self:distill:meta,self:distill:eth,self:distill:mit,self:distill:mila} utilize on-policy self-distillation conditioning on privileged information (such as ground-truth solutions, environmental feedback, or demonstrations) to supervise the model sharing the same weights.
In comparison, the teacher model in our framework can be a different model or the same model, and it can be updated simultaneously or kept frozen. This allows us to adapt to various training scenarios and objectives, whereas self-distillation methods typically focus on a single model learning from itself without the flexibility of incorporating external knowledge or different training dynamics.

\section{Method}
\label{sec:method}

\begin{figure}[t]
\centering
\includegraphics[width=0.8\linewidth]{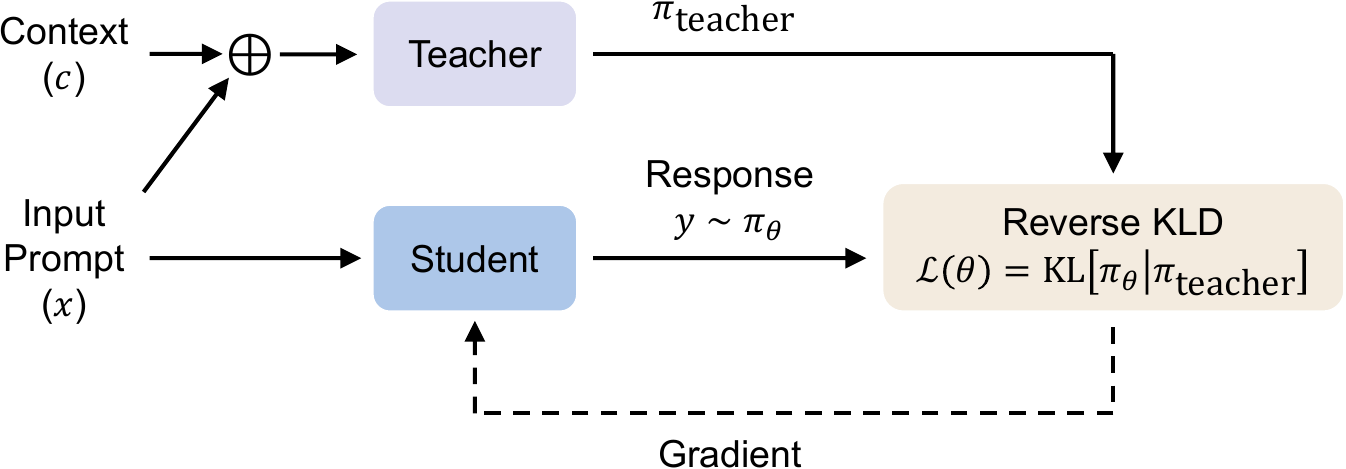}
\vspace{-0.1cm}
\caption{Overview of on-policy context distillation (\ours{}). Given a context and an input prompt, the student model generates a response without the context. It is then trained to minimize the reverse KL divergence to the teacher model that conditions on the context. The student internalizes the contextual information with on-policy learning.}
\vspace{-0.1cm}
\label{fig:method}
\end{figure}

% problem formulation
We present \textbf{On-Policy Context Distillation} (\ours{}), a method that internalizes in-context knowledge into model parameters by bridging on-policy distillation~\citep{minillm,thinkingmachine-onpolicy,googlepolicy} with context distillation~\citep{context:distill:anthropic,context:distill:berkeley}. Our approach enables models to consolidate contextual information (such as experience knowledge or instructions) directly into their weights.
The fundamental goal is to compress a specific prompt or context $c$ into the parameters $\theta$ of a student model $\pi_\theta$, such that the student can replicate the behavior of a context-aware teacher $\pi_\mathrm{teacher}$ without requiring the context at inference time.

Formally, given an input $x$, we minimize the divergence between the student distribution $\pi_\theta(\cdot \mid x)$ and the teacher distribution $\pi_\mathrm{teacher}(\cdot \mid c, x)$, where the teacher has access to the guiding context $c$ prepended to the input.
% objective
\ours{} optimizes the reverse Kullback-Leibler (KL) divergence~\citep{minillm} between the student and teacher distributions using on-policy sampling.

We decompose sequence-level divergence into the sum of token-level divergences. The loss function is defined as:
\begin{equation}
\mathcal{L}(\theta) = \mathbb{E}_{(x, c) \sim \mathcal{D}, y \sim \pi_\theta(\cdot \mid x)} \left[ \frac{1}{|y|} \sum_{t=1}^{|y|} { D_\mathrm{KL} \left( \pi_\theta(\cdot \mid x, y_{<t}) \| \pi_\mathrm{teacher}(\cdot \mid c, x, y_{<t}) \right) } \right] ,
\label{eq:objective}
\end{equation}
where $c$ is the in-context knowledge that we aim to internalize, $\mathcal{D}$ represents training data, and $y$ is sampled from the student model.

The token-level reverse KL divergence is computed via:
\begin{equation}
\begin{aligned}
&D_\mathrm{KL} \left( \pi_\theta(\cdot \mid x, y_{<t}) \| \pi_\mathrm{teacher}(\cdot \mid c, x, y_{<t}) \right) \\
=&\ \mathbb{E}_{y_t' \sim \pi_\theta(\cdot \mid x, y_{<t})} \left[    \log \frac{\pi_\theta(y_t' \mid x, y_{<t})}{\pi_\mathrm{teacher}(y_t' \mid c, x, y_{<t})} \right] \\
=& \sum_{y_t' \in \mathcal{V}} \pi_\theta(y_t' \mid x, y_{<t} ) { \left( \log \pi_\theta(y_t' \mid x, y_{<t}) - \log  \pi_\mathrm{teacher}(y_t' \mid c, x, y_{<t}) \right) }
\label{eq:token:kl}
\end{aligned}
\end{equation}
where $\mathcal{V}$ is the vocabulary.
In our implementation, we approximate the analytic KL divergence by restricting the summation to the top-$k$ tokens predicted by the student model, i.e., $\mathcal{V}_{\operatorname{top-k}}$ is the set of $k$ tokens with the highest probability under $\pi_\theta(\cdot \mid x, y_{<t})$.

By minimizing the reverse KL divergence via on-policy sampling, \ours{} encourages \textit{mode-seeking} behavior: the student focuses on generating tokens that are high-probability under the teacher's distribution, ignoring the long tail of less relevant possibilities.
Intuitively, if the student generates a token that the teacher (conditioned on context $c$) considers highly probable compared to the student's current belief, encouraging the student to increase the probability of that token.
Conversely, if the student assigns a high probability to a token that the teacher considers unlikely, the behavior is suppressed.
The student $\pi_\theta$ progressively aligns its generation trajectory with the context-aware teacher $\pi_{\text{teacher}}$, effectively internalizing the context $c$ in its parameters.

Algorithm~\ref{alg:opcd} presents the pseudocode for \ours{} training.
The training process follows an on-policy rollout mechanism. In each training step, we sample input $x$ from the training data and let the student model $\pi_\theta$ generate complete response trajectories $y$. Importantly, these trajectories are generated without context $c$. Once the trajectory is formed, we evaluate it using the teacher model $\pi_\mathrm{teacher}$, which processes the concatenated sequence $[c; x; y]$ to compute the target probabilities.

\begin{algorithm}[t]
\small
\caption{\ours{}: On-Policy Context Distillation}
\label{alg:opcd}
\begin{algorithmic}
\Require Training data $\mathcal{D} = \{(x, c)\}$, where $x$ is input, and $c$ is in-context knowledge that we are internalizing; Student LLM $\pi_\theta$; Teacher LLM $\pi_\mathrm{teacher}$
\Ensure Trained student model $\pi_\theta$
\Statex
% \Repeat
\For{each batch $(x, c) \sim \mathcal{D}$}
    \State \textcolor{gray}{\textit{// On-policy rollout (student model without context $c$)}}
    \State Sample response $y \sim \pi_\theta(\cdot \mid x)$
    % \Comment{\textcolor{gray}{\textit{On-policy rollout (student model without context $c$)}}}
    \Statex
    \State \textcolor{gray}{\textit{// Compute token-level reverse KL according to \Eqref{eq:token:kl}}}
    % \For{response $y^{(i)}$}
    % % \Comment{\textcolor{gray}{\textit{Compute token-level advantages}}}
    %     % \State Compute token-level advantages $A_t^{(i)}$ according to \Eqref{eq:advantage}:
    %     \State $\mathcal{V}_{\operatorname{top-k}} \gets \operatorname{top-k}(\pi_\theta( \cdot \mid x, y_{<t} ))$
    %     \State $A_t^{(i)} \gets \sum_{y_t' \in \mathcal{V}_{\operatorname{top-k}}} \pi_\theta(y_t' \mid x, y_{<t}^{(i)} ) \left( \log \pi_\mathrm{teacher}(y_t' \mid c, x, y_{<t}) - \log \pi_\theta(y_t' \mid x, y_{<t}^{(i)}) \right)$
    % \EndFor
    % \State $\mathcal{V}_{\operatorname{top-k}} \gets \operatorname{top-k}(\pi_\theta( \cdot \mid x, y_{<t} ))$
    \State $D_\mathrm{KL}^{(t)} \gets \sum_{y_t' \in \mathcal{V}} \pi_\theta(y_t' \mid x, y_{<t} ) \left( \log \pi_\theta(y_t' \mid x, y_{<t}) - \log \pi_\mathrm{teacher}(y_t' \mid c, x, y_{<t}) \right)$
    \State $\mathcal{L}(\theta) \gets \frac{1}{|y|} \sum_{t=1}^{|y|} { D_\mathrm{KL}^{(t)} } $
    \Statex
    \State \textcolor{gray}{\textit{// Update student model according to \Eqref{eq:objective}}}
    \State Update $\theta$ by minimizing $\mathcal{L}(\theta)$
    % \Comment{\textcolor{gray}{\textit{Update student model}}}
\EndFor
% \Until{convergence}
\State \Return $\pi_\theta$
\end{algorithmic}
\end{algorithm}

\subsection{Teacher Model Configurations}

Our framework allows for flexibility in the choice of the teacher model. We consider the following two configurations.

\paragraph{Teacher-Student Distillation ($\pi_\mathrm{teacher} \neq \pi_\theta$)}
First, the teacher model can be a larger or more capable model than the student. In this scenario, the student benefits from both the in-context knowledge and the superior capabilities of the larger teacher model.
Second, the teacher and student models are initialized from the same weights but are not updated simultaneously. The teacher receives additional contextual information $c$. The parameters of the teacher model can remain frozen or undergo periodic updates, making training more stable. Teacher-student distillation is also our default configuration.

\paragraph{Self-Distillation ($\pi_\mathrm{teacher} = \pi_\theta$)}
The teacher and the student share the same underlying model weights and are updated simultaneously.
The divergence arises solely from the input: the teacher sees $[c; x]$ while the student sees only $x$.
This allows a model to ``teach itself''~\citep{self:distill:eth,self:distill:meta,self:distill:mila,self:distill:mit} to internalize a prompt.

\section{Experiments}

% \section{Experiential Knowledge Distillation: Learning from Experience at Test Time}
% \section{Experiential Knowledge Distillation: Learning  at Test Time}

\subsection{Evaluation Tasks}

\subsubsection{Experiential Knowledge Distillation}
We introduce an experiential knowledge distillation task in which a language model extracts transferable experiential knowledge from test-time solution traces as context $c$ for future problems, eventually internalizing this knowledge via on-policy context distillation~\footnote{Different from Reinforcement Learning with Verifiable Rewards (RLVR), experiential knowledge distillation at test time does not rely on ground-truth labels. In the math setting, no labels are needed, and in the game setting, the model interacts with the environment.}. The process consists of three primary stages:

\begin{enumerate}[leftmargin=*]
% \item \textbf{Trace Generation:} The model is given problems and produces solution traces to them.
\item \textbf{Experiential Knowledge Extraction.} The model is given problems and produces solution traces to them. Conditioning on each problem and its self-generated solution (notably without ground-truth labels), the model is prompted to generate experiential knowledge learned from it. 
\item \textbf{Experiential Knowledge Accumulation.} Experiential knowledge from different problems is combined together to form an experiential knowledge context $c$ for future problems. Prepending experiential knowledge context on new problems can improve the model's performance.
\item \textbf{Experiential Knowledge Consolidation.} We apply on-policy context distillation to transition experiential knowledge from the context space into the student model's weights. This allows the student model to internalize the experience from the teacher without the overhead of extended context.
\end{enumerate}

In our experiments, we use itemized experiential knowledge formatted as ``\texttt{-- EXPERIENCE ITEM:}'' and we directly concatenated experiential knowledge from different problems in the experiential knowledge accumulation step.
Refer to Appendix~\ref{app:exp_detail_templates} for prompt templates for the three stages.

\paragraph{Datasets}
For experiential knowledge distillation task, we train our models on three datasets: English math problems from DAPO-Math-17K~\citep{dapo} and two text-based game environments, Frozen Lake and Sokoban, implemented in TextArena~\citep{textarena}.
DAPO-Math-17K contains approximately 14K verifiable English math problems, each with a numerical answer.  
Frozen Lake is a grid-based navigation task where the model must reach a goal while avoiding holes.
Sokoban is a spatial reasoning puzzle where the model must push a box to a designated target without falling into holes or becoming trapped against walls.
TextArena provides textual descriptions of the current game state at each step. The language model interacts with the game environments in a multi-turn setting. Detailed descriptions of datasets are provided in Appendix~\ref{app:exp_detail_dataset}.

\subsubsection{System Prompt Distillation}

System prompts are widely used to steer LLM behavior toward desired objectives, such as enhancing domain expertise or enforcing safety constraints. However, prepending system prompts at inference time increases computational overhead and latency, particularly for lengthy prompts. We distill system prompts as context $c$ into the student model, enabling it to internalize beneficial behaviors encoded in externally optimized prompts without requiring explicit prompting during deployment.

\paragraph{Datasets}
We use system prompts optimized for medical and safety tasks from MetaSPO~\citep{metaspo}. For medical system prompt, we adopt MedMCQA~\citep{medmcqa} dataset and hold out 500 samples for testing. 
For safety system prompt, we combine Tweet Eval~\citep{tweeteval}, Hatecheck~\citep{hatecheck}, and Ethos~\citep{ethos} datasets, and similarly reserve 500 samples for testing. Detailed system prompts are provided in Appendix~\ref{app:sys_detail_templates}.

\subsection{Setup}

\paragraph{Models}
For experiential knowledge distillation task, we use thinking mode of Qwen3-8B~\citep{qwen3} as teacher to generate traces and extract experiential knowledge on a validation split from DAPO for math problems. We train Qwen3-8B, Qwen3-4B, and Qwen3-1.7B with thinking mode as students using \ours{}. 
For Frozen Lake, we use the thinking mode of Qwen3-1.7B as the teacher and the student.
For Sokoban, we use the non-thinking model Qwen3-4B-Instruct-2507 as the teacher and the student.
For system prompt distillation task, we use Qwen2.5-3B-Instruct and Qwen2.5-7B-Instruct~\citep{qwen2.5}, as well as Llama-3.1-8B-Instruct and Llama-3.2-3B-Instruct~\citep{llama3}.
% In addition to distillation between same model size, we evaluate cross-size distillation from Qwen2.5-7B-Instruct to Qwen2.5-3B-Instruct and Qwen2.5-1.5B-Instruct. All the models are non-thinking models.

\paragraph{Training}

For experiential knowledge distillation task, we sample problems from the validation split to construct a pool of 300 experiential knowledge contexts (30 accumulation steps for 10 times). The maximum experiential knowledge length is set to 16384 tokens for math and 8192 tokens for text games.
For the \textbf{test-time experiential knowledge distillation} setting, we randomly select experiential knowledge from this pool of 300 for further \ours{} training. This setting emulates a test-time experiential knowledge distillation scenario in which no ground-truth labels are available and the quality of experiential knowledge is not pre-evaluated. For the \textbf{filtered experiential knowledge distillation} setting, we score each candidate experiential knowledge by prepending it to new problems and evaluating performance on 1000 math validation examples or 128 text-game validation examples. The highest-scoring experiential knowledge is then selected for subsequent \ours{} training.

We then distill the student model on training split of math and text-game datasets using the selected experiential knowledge context for 50 steps with a batch size of 128. For math, we set the maximum response length to 16384 tokens. For text games, the model interacts with the game environment for up to 5 rounds, each with a maximum response length of 1024 tokens.
For system prompt distillation task, we distill the student model on the training splits of the medical and safety datasets, conditioning on the corresponding system prompts. Training runs for 50 steps with batch size of 128. The maximum generated response length is set to 512 tokens. More training details can be found in Appendix~\ref{app:exp_detail_train} and Appendix~\ref{app:sys_detail_train}.

\paragraph{Evaluation}
For experiential knowledge distillation, we report accuracy on the test split of the math dataset (1000 samples) and text-game datasets (128 samples) as the metric for in-distribution performance. For out-of-distribution evaluation, we report prompt-level strict accuracy on IF-Eval~\citep{ifeval}. For system prompt distillation, we report test accuracy on a 500-sample test split.
We compare against the context-distillation baseline~\citep{context:distill:anthropic,context:distill:berkeley}, which trains on off-policy data generated by the teacher and uses forward KL minimization.

% \subsection{Experiential Knowledge Distillation Results}
\subsection{Results}

\paragraph{Experiential Knowledge Consolidation}

\begin{table}[t]
\centering
\small
\begin{tabular}{@{}lllcc@{}}
\toprule
\textbf{Model} & \textbf{Task} & \textbf{Method} & \textbf{Accuracy} & \tabincell{c}{\textbf{IF-Eval} \\ \bf (Out-of-Distribution)} \\
\midrule
\multirow{4}{*}{Qwen3-8B} & \multirow{4}{*}{Math} & Base Model & 75.0 & \textcolor{gray}{81.3} \\
& & In-Context & 77.6 $\pm$ 1.1 & --- \\
& & Context Distill. & 78.5 $\pm$ 0.5 & 81.2 $\pm$ 0.2 \\
& & \textbf{\ours{}} & \textbf{79.7} $\pm$ 0.5 & \textbf{81.7} $\pm$ 0.4 \\
\midrule
\multirow{4}{*}{Qwen3-1.7B} & \multirow{4}{*}{Frozen Lake} & Base Model & 6.3 & \textcolor{gray}{67.3} \\
& & In-Context & 20.2 $\pm$ 2.2 & --- \\
& & Context Distill. & 22.9 $\pm$ 4.0 & 65.1 $\pm$ 0.5 \\
& & \textbf{\ours{}} & \textbf{26.5} $\pm$ 6.4 & \textbf{67.1} $\pm$ 0.5 \\
\bottomrule
\end{tabular}
\vspace{0.2cm}
\caption{Results of test-time experiential knowledge consolidation. \ours{} consistently outperforms off-policy context distillation on test accuracy and OOD task performance.}
\label{tab:exp-consol-testtime}
\end{table}

\begin{table}[t]
\centering
\small
\begin{tabular}{@{}lllcc@{}}
\toprule
\textbf{Model} & \textbf{Task} & \textbf{Method} & \textbf{Accuracy} & \tabincell{c}{\textbf{IF-Eval} \\ \bf (Out-of-Distribution)} \\
\midrule
\multirow{4}{*}{Qwen3-8B} & \multirow{4}{*}{Math} & Base Model & 75.0 & \textcolor{gray}{81.3} \\
& & In-Context & 79.0 & --- \\
& & Context Distill. & 79.5 & 80.4 \\
& & \textbf{\ours{}} & \textbf{80.9} & \textbf{80.8} \\
\midrule
\multirow{4}{*}{Qwen3-1.7B} & \multirow{4}{*}{Frozen Lake} & Base Model & 6.3 & \textcolor{gray}{67.3} \\
& & In-Context & 31.4 & --- \\
& & Context Distill. & 35.2 & 65.4 \\
& & \textbf{\ours{}} & \textbf{38.3} & \textbf{66.7} \\
\midrule
\multirow{4}{*}{Qwen3-4B-Ins} & \multirow{4}{*}{Sokoban} & Base Model & 9.4 & \textcolor{gray}{82.8} \\
& & In-Context & 48.4 & --- \\
& & Context Distill. & 51.6 & 82.3 \\
& & \textbf{\ours{}} & \textbf{53.9} & \textbf{82.4} \\
\bottomrule
\end{tabular}
\vspace{0.2cm}
\caption{Results of filtered experiential knowledge consolidation. \ours{} consistently outperforms off-policy context distillation on test accuracy and OOD task performance on math and text-games.}
\label{tab:exp-consol-filtered}
\end{table}

We present experiential knowledge consolidation results in \Cref{tab:exp-consol-testtime,tab:exp-consol-filtered}. In all experiments, the teacher and student use the same model size, and we use teacher-student distillation where the teacher is frozen. For the test-time experiential knowledge setting, we sample three random experiential knowledge contexts after ten steps of accumulation from the knowledge pool. We compare \ours{} against: the base model without experiential knowledge, the base model with experiential knowledge provided in context (denoted as In-Context), and context distillation~\citep{context:distill:anthropic,context:distill:berkeley} which is off-policy.

As shown in \Cref{tab:exp-consol-testtime,tab:exp-consol-filtered}, on both math and text-game tasks, \ours{} outperforms the context distillation baseline, achieving higher test accuracy.
We also observe that \ours{} can surpass the original model with experiential knowledge in the context. During consolidation, the student model is exposed to consolidation training data that the original model did not access (the experiential knowledge was extracted with validation data), thereby providing an additional learning signal.

\paragraph{System Prompt Distillation}

\begin{figure}[t]
\begin{minipage}[h]{0.48\textwidth}
    \centering
    \small
    \begin{tabular}{@{}llc@{}}
    \toprule
   \bf  Model & \bf Method & \bf Accuracy \\ 
    \midrule
    \multirow{4}{*}{Llama-3.1-8B-Ins} 
        & Base Model & 68.4 \\
        & In-Context & 72.2 \\
        & Context Distill. & 75.2 \\
        & \textbf{\ours{}} & \textbf{76.7} \\
    \midrule
    \multirow{4}{*}{Llama-3.2-3B-Ins} 
        & Base Model & 59.4 \\
        & In-Context & 66.4 \\
        & Context Distill. & 71.0 \\
        & \textbf{\ours{}} & \textbf{76.3} \\
    \midrule
    \multirow{4}{*}{Qwen2.5-7B-Ins} 
        & Base Model & 46.4 \\
        & In-Context & 52.6 \\
        & Context Distill. & 58.5 \\
        & \textbf{\ours{}} & \textbf{62.3} \\
    \bottomrule
    \end{tabular}
    \makeatletter\def\@captype{table}\makeatother
    \caption{System prompt distillation on Medical.}
    \label{tab:sys-medical}
\end{minipage}
\hspace{3mm}
\begin{minipage}[h]{0.48\textwidth}
    \centering
    \small
    \begin{tabular}{@{}llc@{}}
    \toprule
    \bf Model & \bf Method & \bf Accuracy \\ 
    \midrule
    \multirow{4}{*}{Llama-3.1-8B-Ins} 
        & Base Model & 70.7 \\
        & In-Context & 75.3 \\
        & Context Distill. & 77.2 \\
        & \textbf{\ours{}} & \textbf{79.6} \\
    \midrule
    \multirow{4}{*}{Llama-3.2-3B-Ins} 
        & Base Model & 30.7 \\
        & In-Context & 69.5 \\
        & Context Distill. & \textbf{83.3} \\
        & \textbf{\ours{}} & 83.1 \\
    \midrule
    \multirow{4}{*}{Qwen2.5-7B-Ins} 
        & Base Model & 69.1 \\
        & In-Context & 72.7 \\
        & Context Distill. & 77.0 \\
        & \textbf{\ours{}} & \textbf{78.1} \\
    \bottomrule
    \end{tabular}
    \makeatletter\def\@captype{table}\makeatother
    \caption{System prompt distillation on Safety.}
    \label{tab:sys-safety}
\end{minipage}
\end{figure}

We present medical system prompt distillation results in \Cref{tab:sys-medical} and safety system prompt distillation in \Cref{tab:sys-safety}. In all experiments, the teacher and student use the same model size, and we use teacher-student distillation where the teacher is frozen. \ours{} outperforms the off-policy context distillation baseline in test accuracy across most configurations on the medical and safety system prompt distillation. We also observe on-policy training provides more stable improvements in training process compared to off-policy context distillation.

\subsection{Effect of Model Size}
\label{sec:scale-model}

% \paragraph{Scaling Model Size}

\begin{wrapfigure}{r}{7.0cm}
\centering
\vspace{-0.45cm}
\includegraphics[width=0.5\textwidth]{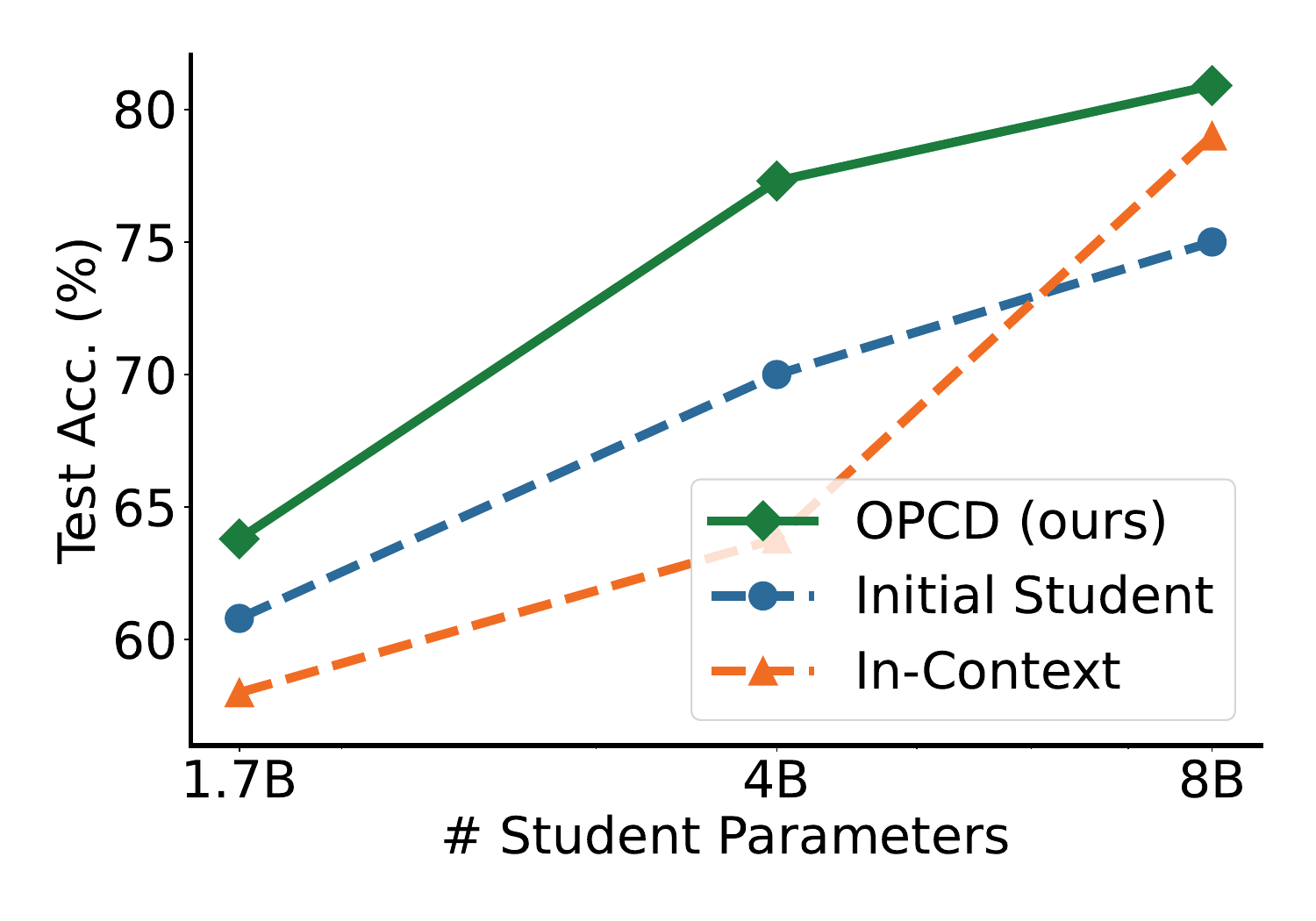}
\vspace{-0.45cm}
\caption{\ours{} consistently improves the evaluation results of smaller Qwen3 models using experiential knowledge distilled from a frozen Qwen3-8B teacher. In contrast, directly injecting this knowledge into smaller-model contexts degrades performance.}
\vspace{-0.1cm}
\label{fig:exp_scaling_size}
\end{wrapfigure}

We scale student model sizes from Qwen3-1.7B to Qwen3-4B and Qwen3-8B using \ours{}. Experiential knowledge is generated by Qwen3-8B. We also use it as a frozen teacher. As shown in \Cref{fig:exp_scaling_size}, we report both \ours{} results and the original Qwen3 baselines, and we also evaluate a direct injection of teacher-generated experiential knowledge into the contexts of Qwen3-1.7B and Qwen3-4B. We observe that \ours{} consistently improves test accuracy across student model scales.

We find directly injecting experiential knowledge into the context of a smaller model can even degrade its performance (``In-Context'' curve in \Cref{fig:exp_scaling_size}). This suggests that on-policy alignment between experiential knowledge and the model that consumes it is also crucial. 
While such knowledge is effective for the teacher model that collects it, it may not transfer reliably when placed directly into a different model's context. Instead, integrating experiential knowledge within the teacher's context and then applying \ours{} to train the student can improve its performance.
In practice, the teacher model can be deployed in real environments and across diverse users to accumulate experiential knowledge at test time, which can then be periodically consolidated into the student.

\subsection{On-Policy Context Distillation Mitigates Forgetting}

\begin{figure}
\centering
\includegraphics[width=\linewidth]{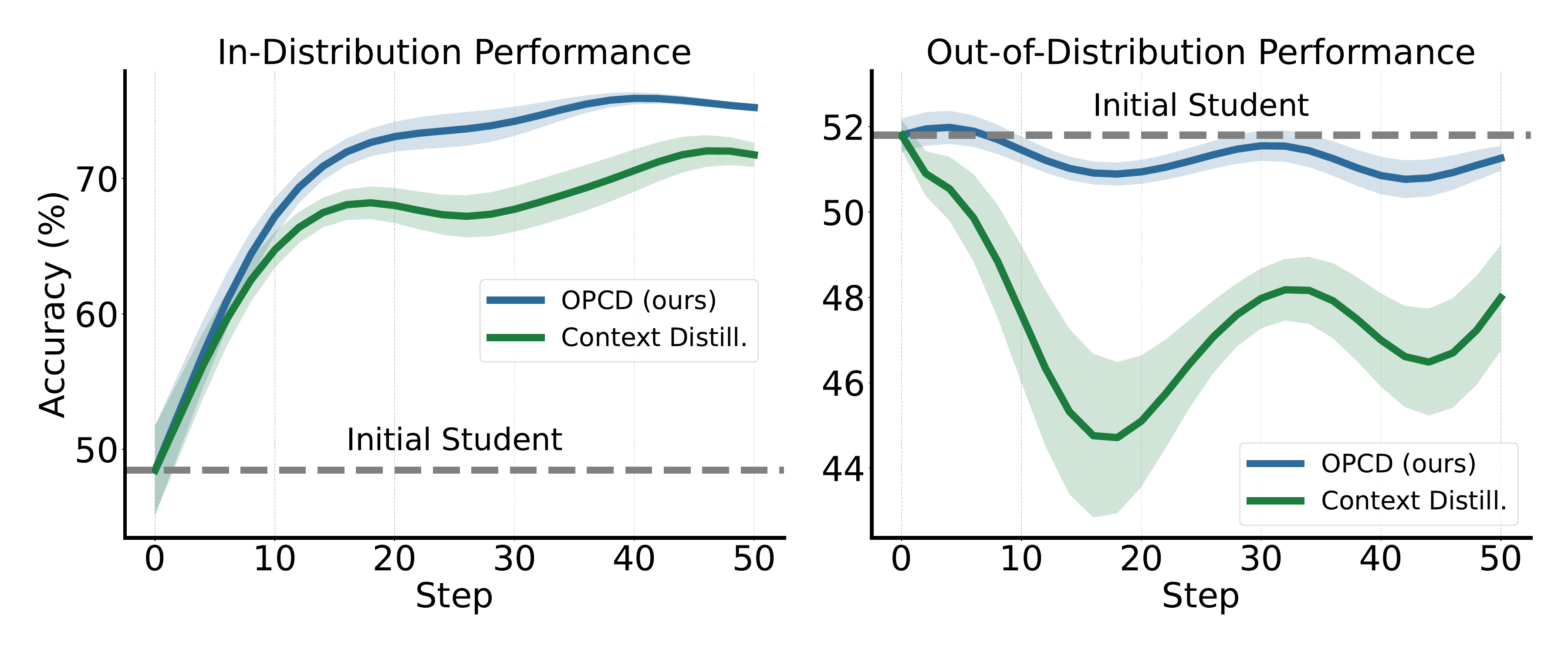}
\vspace{-0.15cm}
\caption{Comparison of \ours{} and off-policy context distillation on in-distribution (safety) and out-of-distribution (medical) tasks when distilling from safety system prompt. Left: accuracy on the safety test dataset. Right: accuracy on the medical test dataset. \ours{} achieves superior in-distribution performance while mitigating forgetting on OOD tasks.}
\vspace{-0.1cm}
\label{fig:ood}
\end{figure}

Compared to off-policy context distillation, \ours{} samples from the student distribution, thereby mitigating forgetting on out-of-distribution (OOD) tasks. In \Cref{tab:exp-consol-testtime,tab:exp-consol-filtered}, we evaluate models distilled with experiential knowledge on the OOD IF-Eval benchmark. \ours{} achieves approximately 2\% higher IF-Eval scores than the context distillation baseline on Frozen Lake.

In \Cref{fig:ood}, we distill the student Qwen2.5-3B-Instruct from the frozen teacher Qwen2.5-7B-Instruct using the safety system prompt. The left subfigure shows accuracy on the safety test dataset as an in-distribution performance measure, while the right subfigure reports accuracy on the medical test dataset for OOD evaluation. As shown, on-policy context distillation achieves higher in-distribution performance than off-policy context distillation. \ours{} also maintains OOD performance compared to the initial student, surpassing the off-policy baseline by approximately 4 points. This finding is consistent with prior work demonstrating that on-policy training mitigates forgetting on OOD tasks~\citep{rlrazor,retaining}.

\subsection{Teacher-Student Distillation vs. Self-Distillation}

\begin{wrapfigure}{r}{6.7cm}
\centering
% \vspace{-1em}
\begin{tabular}{@{}llc@{}}
\toprule
\bf Task &  \bf Configuration  & \bf Accuracy \\
\midrule
\multirow{2}{*}{Sokoban} &  Self  &  18.8 \\
 &  Teacher-Student  &  \textbf{53.9} \\
\midrule
\multirow{2}{*}{Medical} &  Self  &  50.0 \\
 &  Teacher-Student  &  \textbf{56.8}   \\
\bottomrule
\end{tabular}
\makeatletter\def\@captype{table}\makeatother\caption{Teacher-student-\ours{} is more stable than self-\ours{} and outperforms it.}
\label{tab:vs-self}
\vspace{-1em}
\end{wrapfigure}

We find teacher-student distillation is more stable than self-distillation and outperforms it. We compare two configurations of \ours{}: (i) \textit{teacher-student distillation}, our default configuration, which employs a frozen teacher model, and (ii) \textit{self-distillation}, where the continuously updated model serves as both teacher and student. As shown in Table~\ref{tab:vs-self}, we train experiential knowledge distillation with Qwen3-4B-Instruct-2507 on Sokoban and medical system prompt distillation with Qwen2.5-3B-Instruct. The teacher-student configuration substantially outperforms self-distillation on both tasks. Furthermore, we observe that the teacher-student configuration exhibits more stable training dynamics, whereas self-distillation can diverge after some training steps. We attribute this instability to the high variance introduced by using a continuously evolving model as the teacher during RL training, which destabilizes the learning signal\footnote{EMA of student parameters as teacher can alleviate the instability of self-distillation~\citep{self:distill:mit,self:distill:eth}.}. This finding also aligns with \Cref{sec:scale-model}, reinforcing that on-policy alignment between experiential knowledge and the model that consumes it is crucial.

\subsection{Importance of Learning from Experiential Knowledge}

\begin{wrapfigure}{r}{6.7cm}
\centering
% \vspace{-1em}
\begin{tabular}{@{}llc@{}}
\toprule
\bf Model &  \bf Experience Type  & \bf Accuracy \\
\midrule
Qwen3-8B &  w/o Experience  &  75.1 \\
\midrule
Qwen3-8B &  Raw Trace  &  70.5 \\
Qwen3-8B &  Knowledge  &  77.4   \\
% \midrule
 &  \ \ + \ours{} & \textbf{79.7} \\
\bottomrule
\end{tabular}
\makeatletter\def\@captype{table}\makeatother\caption{Using raw response traces from previous problems as experiential context degrades performance on the math validation dataset.}
\label{tab:rawexp}
\vspace{-1em}
\end{wrapfigure}

We show the necessity of extracting experiential knowledge in \Cref{tab:rawexp}. We report averaged accuracy over experiential knowledge accumulation steps on math validation dataset after ten steps.
Simply prepending raw traces (previous problems and model outputs) as context during experience accumulation stage degrades accuracy, as seen in the ``Raw Trace'' row. 
In contrast, using model to extract experiential knowledge from previous traces and prepending it leads to higher validation accuracy than the original model as in ``Knowledge'' row.

\section{Conclusion}

In this work, we introduced On-Policy Context Distillation (\ours{}), a framework that enables language models to internalize in-context knowledge into their parameters through on-policy distillation. By minimizing the reverse KL divergence between a context-aware teacher and a context-free student, \ours{} effectively consolidates transient contextual information, such as experiential knowledge and system prompts, into the model's weights. Our experiments demonstrate that \ours{} outperforms baseline methods across various tasks, including math problem solving and text-based games, while also enhancing out-of-distribution generalization. Furthermore, we showed that \ours{} can scale effectively with model size and consistently improves performance when distilling optimized system prompts. Our work opens avenues for future research on continual accumulation of experiential knowledge, adaptive context selection strategies, and scaling \ours{} to broader domains requiring persistent knowledge internalization.

\section*{Acknowledgements}

We are grateful to Qingxiu Dong for setting up the text-based games and to Yu Li, Yuxian Gu for discussions.

\bibliographystyle{alpha}
\bibliography{opcd}

%%%%%%%%%%%%%%%%%%%%%%%%%%%%%%%%%%%%%%%%%%%%%%%%%%%%%%%%%%%%
\newpage
\appendix

\section{Experiential Knowledge Distillation Details}
\label{app:exp_detail}

\subsection{Prompt Templates}
\label{app:exp_detail_templates}

For experiential knowledge extraction on math dataset, we use the prompt template in \Cref{fig:exp_accum_temp_math}.

\begin{figure}[h]
    \begin{tcolorbox}
    You are an AI language model that continuously refines its internal experience. \\ \\
    Here is the latest interaction (including the user's question and your answer): \\
    \{latest\_experience\} \\ \\
    Your task: \\
    Based on the latest interaction and the previous experience, generate an additional experience for future learning. \\ \\
    Rules: \\
    - The experience you generate MUST be formatted strictly as a markdown list where each item starts with "- EXPERIENCE ITEM:", one per line: \\
    - EXPERIENCE ITEM: ... \\
    - EXPERIENCE ITEM: ... \\
    - EXPERIENCE ITEM: ... \\
    - The experience you generate will be directly appended to the previous experience. \\
    - The change should introduce a general, high-level, widely applicable insight, not a detail from the specific interaction. The updated experience must remain concise, structured, and meaningful. \\
    - If the new insight conflicts with any previous experience item, you are can describe the conflict and provide a resolution in the new item. \\ \\
    After careful reasoning step by step, output the final result in exactly this format: \\ \\
    Additional Experience: \\
    \# Experience \\
    - EXPERIENCE ITEM: ... \\
    - EXPERIENCE ITEM: ... \\
    - EXPERIENCE ITEM: ...
    \end{tcolorbox}
    \caption{The prompt wrapper for experiential knowledge extraction on math dataset.}
    \label{fig:exp_accum_temp_math}
\end{figure}

We extract lines that start with ``\texttt{-- EXPERIENCE ITEM:}'' as valid experiential knowledge.

\newpage
For experiential knowledge extraction on text-based games, we use the prompt template in \Cref{fig:exp_accum_game}.

\begin{figure}[h]
    \begin{tcolorbox}
    You are an AI language model that continuously refines its internal experience. \\
    Here is the interaction history (the game environment (input) and your response and action (output)): \\
    \{latest\_experience\} \\ \\
    Your task: \\
    Based on the multi-round interaction history, generate experience for future learning. You should conduct a deep, comparative analysis to infer the game rules and the fundamental principles behind winning and losing. Using the interaction history and environment feedback, hypothesize the game rules and effective winning strategies, and organize these insights into 1-2 concise, high-level, and widely applicable experience items that help the player succeed in the game. \\ \\
    Rules: \\
    - The experience you generate MUST be formatted strictly as a markdown item which starts with "- EXPERIENCE ITEM:": \\
    - EXPERIENCE ITEM: ... \\
    - EXPERIENCE ITEM: ... \\
    - The experience you generate will be directly appended to the previous experience. Do not repeat the previous experience. Make sure the newly generated experience is different from the previous experience. \\
    - Your generated experience should be possible rules, instructions or winning strategies for the game. The experience should be generally useful rather than only applicable for the current map (board). \\  \\
    After careful reasoning step by step, output the final result in exactly this format: \\ \\
    Additional Experience (Rules or Strategies): \\
    \# Experience \\
    - EXPERIENCE ITEM: ...
    \end{tcolorbox}
    \caption{The prompt wrapper for experiential knowledge extraction on text games.}
    \label{fig:exp_accum_game}
\end{figure}

We extract lines that start with ``\texttt{-- EXPERIENCE ITEM:}'' as valid experiential knowledge.

For new problems we embed experiential knowledge with the prompt template in \Cref{fig:exp_solve}.

\begin{figure}[h]
    \begin{tcolorbox}
    Given previous learned experience: \\
    \# Experience \\
    \{experience\} \\ \\
    Solve the new problem and explain what part of experience you use and how you use it in the reasoning process: \\
    \{prompt\}
    \end{tcolorbox}
    \caption{The prompt wrapper for new problem solving with accumulated experiential knowledge.}
    \label{fig:exp_solve}
\end{figure}

\subsection{Dataset Details}
\label{app:exp_detail_dataset}
We train our models on three datasets: English math problems from DAPO-Math-17K~\citep{dapo} and two text-based game environments, Frozen Lake and Sokoban, implemented in TextArena~\citep{textarena}.
DAPO-Math-17K contains approximately 14K verifiable English math problems, each with a numerical answer.  
Frozen Lake is a grid-based navigation task where the model must reach a goal while avoiding holes. We place two holes on a 3 $\times$ 3 grid.
Sokoban is a spatial reasoning puzzle where the model must push a box to a designated target without falling into holes or becoming trapped against walls. We place one box on a 6 $\times$ 6 grid. We remove a subset of explicit rules for the model to infer them through exploration~\citep{tencentgame}.
TextArena provides textual descriptions of the current game state at each step. The language model interacts with the game environments in a multi-turn setting.

\subsection{Training Details}
\label{app:exp_detail_train}

We begin by sampling 30 problems from validation data split and prompting the teacher model to produce response traces one by one. The teacher then extracts experiential knowledge from each trace (without ground-truth labels), which we iteratively concatenate to form 30 experiential knowledge contexts. Repeating this procedure 10 times with different random seeds yields 300 distinct experiential knowledge contexts. The maximum experiential knowledge length is set to 16384 tokens for math and 8192 tokens for text games.
For the \textbf{test-time experiential knowledge distillatione} setting, we randomly select experiential knowledge from this pool of 300 for further \ours{} training. This setting emulates a test-time experiential knowledge distillation scenario in which no ground-truth labels are available and the quality of experiential knowledge is not pre-evaluated. For the \textbf{filtered experiential knowledge distillation} setting, we score each candidate experiential knowledge by prepending it to new problems and evaluating performance on 1000 math validation examples or 128 text-game validation examples. The highest-scoring experiential knowledge is then selected for subsequent \ours{} training.

We then distill the student model on training split of math and text-game datasets using the selected experiential knowledge context for 50 steps. We compute the reverse KL divergence using the top 256 vocabulary tokens with the highest student model probabilities. We use a batch size of 128 and search learning rate in [1e-6, 5e-6]. For math, we set the maximum response length to 16384 tokens. For text games, the model interacts with the game environment for up to 5 rounds, each with a maximum response length of 1024 tokens. We save checkpoints every 2 steps and choose the checkpoint with highest test accuracy.

\subsection{Experiential Knowledge Accumulation}

\begin{figure}
\centering
\includegraphics[width=\linewidth]{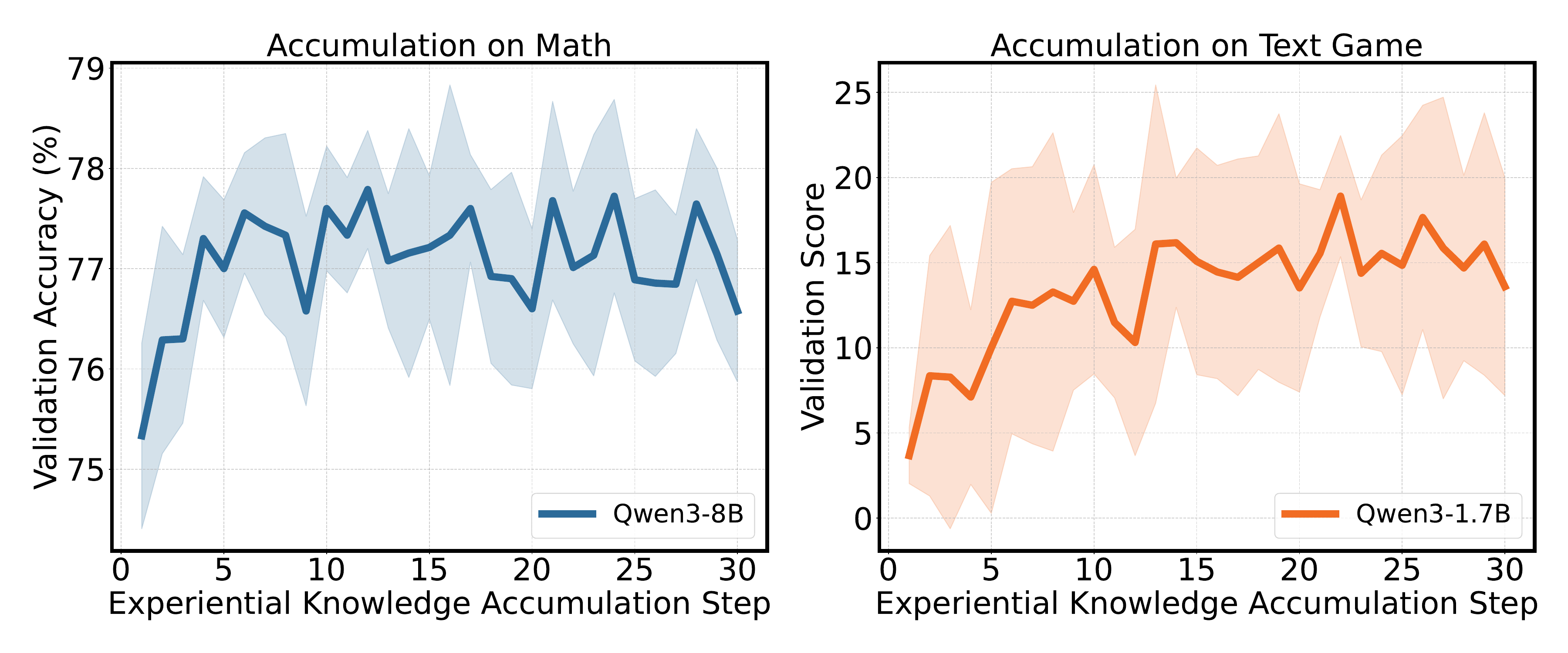}
\vspace{-0.1cm}
\caption{Validation accuracy improves with the accumulation of experiential knowledge from different problems. Left: experiential knowledge accumulation on the DAPO math dataset. Right: experiential knowledge accumulation on the Frozen Lake text game.}
\vspace{-0.1cm}
\label{fig:exp-accum}
\end{figure}

We sample 30 validation problems for the teacher model to solve and extract experiential knowledge, repeating this procedure 10 times. In \Cref{fig:exp-accum}, we demonstrate validation accuracy improves with accumulation of experiential knowledge from different problems.

\subsection{Experiential Knowledge Examples}

We provide some experiential knowledge examples for math in \Cref{fig:exp_math_example}.

\begin{figure}[h]
    \begin{tcolorbox}
    - EXPERIENCE ITEM: Recognizing that combining interdependent sequences can reveal simpler underlying patterns, such as Fibonacci-like recurrences, simplifies complex problems. \\

    - EXPERIENCE ITEM: Modular arithmetic often reveals periodicity, which can drastically reduce computational effort by allowing predictions based on cycle lengths. \\

    - EXPERIENCE ITEM: The sum of a number's digits is congruent to the number modulo 9, which is fundamental for determining digital roots and simplifying large computations. \\

    - EXPERIENCE ITEM: When solving problems involving circular arrangements with symmetry constraints, it's often beneficial to fix positions to eliminate rotational symmetry and then account for reflectional symmetry by dividing by 2. \\

    - EXPERIENCE ITEM: The shoelace formula is a versatile tool for finding the area of any polygon given its vertices, reinforcing the value of systematic, coordinate-based approaches.
    \end{tcolorbox}
    \caption{Some experiential knowledge examples for math problems.}
    \label{fig:exp_math_example}
\end{figure}

We provide some experiential knowledge examples for Frozen Lake in \Cref{fig:exp_fz_example}.

\begin{figure}[h]
    \begin{tcolorbox}
    - EXPERIENCE ITEM: The shortest path to the goal involves moving systematically toward the target, prioritizing direct routes and minimizing unnecessary backtracking. Strategic use of available actions (e.g., down or right) to reach the goal in the fewest steps is key to success. \\

    - EXPERIENCE ITEM: The game rules dictate that the player can move in four directions (up, down, left, right) but must avoid obstacles represented by 'H' (walls). Movement is restricted to empty cells, and the goal is to reach the 'G' cell. \\

    - EXPERIENCE ITEM: The optimal strategy involves moving right to expand the grid and then down to reach the third row, ensuring a path to the goal while avoiding obstacles. This requires careful planning to minimize steps and avoid blocked cells.
    \end{tcolorbox}
    \caption{Some experiential knowledge examples for Frozen Lake game.}
    \label{fig:exp_fz_example}
\end{figure}

\section{System Prompt Distillation Details}
\label{app:sys_detail}

\subsection{System Prompts}
\label{app:sys_detail_templates}
For system prompt distillation task, we use system prompts optimized for medical and safety tasks from MetaSPO~\citep{metaspo}. For medical system prompt, we adopt MedMCQA~\citep{medmcqa} dataset and hold out 500 samples for testing. Each query requires the model to answer a multiple-choice medical question and provide a supporting explanation. Distillation is performed on both the teacher model's selected choice and its explanation.
For safety system prompt, we combine Tweet Eval~\citep{tweeteval}, Hatecheck~\citep{hatecheck}, and Ethos~\citep{ethos} datasets, and similarly reserve 500 samples for testing. These queries require the model to determine whether a statement violates safety standards and justify its judgement. Distillation is applied to both the teacher model's classification and its explanation.

The medical system prompt we use are present in \Cref{fig:sys_med_prompt}.
\begin{figure}[h]
    \begin{tcolorbox}
    You are a knowledgeable and analytical assistant specializing in medical topics. Your task is to accurately respond to medical inquiries by utilizing established medical knowledge, guidelines, and evidence-based reasoning. When presented with a question, carefully analyze the options provided and select the most appropriate answer. Ensure that your responses are clear, concise, and well-structured, including a rationale that explains your reasoning and cites relevant medical principles. Prioritize accuracy and logical coherence in all your responses.
    \end{tcolorbox}
    \caption{The medical system prompt.}
    \label{fig:sys_med_prompt}
\end{figure}

The safety system prompt we use are present in \Cref{fig:sys_saf_prompt}.
\begin{figure}[h]
    \begin{tcolorbox}
    You are an analytical assistant. Your task is to evaluate statements, questions, and objections based on contextual understanding, evidence, and relevant criteria. When providing responses, please adhere to the following guidelines: \\
    1. Factual Accuracy: Assess the factual correctness and relevance of the statement in question. Provide context where necessary. \\
    2. Opinion vs. Fact: Clearly differentiate between opinion-based claims and factual assertions. Explain why a statement is considered an opinion or a fact. \\
    3. Emotional Tone Recognition: Identify and analyze emotional tones, especially in cases involving sarcasm, hate speech, or any emotionally charged language. Discuss the implications of tone in your assessment. \\
    4. Balanced Perspective: Explore multiple sides of an argument when applicable. Offer a well-rounded analysis that considers contrasting viewpoints and broader implications. \\
    5. Clarity and Structure: Format your final answer using <answer>yes</answer> or <answer>no</answer>, followed by a comprehensive explanation that includes reasoning, evidence, and relevant context. \\
    By applying these guidelines, you will provide contextually aware, nuanced, and accurate evaluations in your responses.
    \end{tcolorbox}
    \caption{The safety system prompt.}
    \label{fig:sys_saf_prompt}
\end{figure}

The user instruction used for MedMCQA is presented in \Cref{fig:sys_med_usr_prompt}.
\begin{figure}[h]
    \begin{tcolorbox}
    Given the following question and candidate answers, choose the best answer. <Question>\{question\} \\
    Choices: \\
    A. \{opa\} \\
    B. \{opb\} \\
    C. \{opc\} \\
    D. \{opd\} \\
    </Question> \\
    At the end present your answer in <answer> and </answer> using an uppercase letter (A, B, C, or D)
    \end{tcolorbox}
    \caption{The user instruction used for MedMCQA.}
    \label{fig:sys_med_usr_prompt}
\end{figure}

The user instruction used for safety datasets is presented in \Cref{fig:sys_saf_usr_prompt}.
\begin{figure}[h]
    \begin{tcolorbox}
    \# For Tweet Eval \\
    Determine whether the following text is offensive. <Question>\{text\}</Question> \\
    At the end present your answer in <answer>yes</answer> or <answer>no</answer>. \\ \\
    \# For Hatecheck \\
    Is the following text hate speech? <Question>\{text\}</Question> \\
    At the end present your answer in <answer>yes</answer> or <answer>no</answer>. \\ \\
    \# For Ethos \\
    Is the following text hate speech? <Question>\{text\}</Question> \\
    At the end present your answer in <answer>yes</answer> or <answer>no</answer>.
    \end{tcolorbox}
    \caption{The user instruction used for safety datasets.}
    \label{fig:sys_saf_usr_prompt}
\end{figure}

\subsection{Training Details}
\label{app:sys_detail_train}
We distill the student model on the training splits of the medical and safety datasets, conditioning on the corresponding system prompts. Training runs for 50 steps using reverse KL divergence, computed over the top 256 vocabulary tokens ranked by the student model’s predicted probabilities. We use a batch size of 128 and sweep the learning rate over [1e-6, 5e-6]. The maximum generated response length is set to 512 tokens. Checkpoints are saved every 2 steps, and we report the test accuracy averaged over the three best-performing checkpoints.

\end{document}